\documentclass[lnbip]{svmultln}
\usepackage{CJKutf8}

\usepackage{makeidx}  
\usepackage{amssymb}
\usepackage{nccmath, mathtools}

\usepackage[utf8]{inputenc}
  \usepackage{savesym}
  \usepackage{mathptmx}
  \savesymbol{hbar}
  \usepackage{helvet}
  \usepackage{courier}
  \usepackage{type1cm}           
  \usepackage{fouriernc}
  \usepackage{esvect}   
  \usepackage{chngcntr}   
  \renewcommand{\thechapter}{\Roman{chapter}}
  \renewcommand{\thesection}{\arabic{section}}

\spnewtheorem*{remark}{Remark}{\itshape}{\rmfamily}
\spnewtheorem{theo}{Theorem}[section]{\bfseries}{\itshape}
\spnewtheorem{lemm}[theo]{Lemma}{\bfseries}{\itshape}
\spnewtheorem{defn}[theo]{Definition}{\bfseries}{\itshape}

\usepackage{titlesec}
\titleformat{\section}
  {\normalfont\large\bfseries}{\thesection.}{0.4em}{}

  \counterwithin{definition}{section}
  \counterwithin{theorem}{section}
  \counterwithin{lemma}{section}

\usepackage{graphicx}
\usepackage{subfigure}

\usepackage{threeparttable}
\usepackage{multirow}
\usepackage{multicol}

\usepackage{makecell}

\begin{document}
\begin{sloppypar}

\begin{CJK*}{UTF8}{gbsn}

\mainmatter              
\title{Generalized W-Net:\\Arbitrary-style Chinese Character Synthesization}

\titlerunning{Generalized W-Net: Arbitrary-style Chinese Character Synthesization}  
%
\author{Haochuan Jiang\inst{1} \and Guanyu Yang\inst{2} \and Fei Cheng\inst{3}   \and Kaizhu Huang\inst{4}
\thanks{This research is funded by XJTLU Research Development Funding 20-02-60. Computational resources utilized in this research are provided by the School of Robotics, XJTLU Entrepreneur College (Taicang), and the School of Advanced Technology, Xi’an Jiaotong-Liverpool University.}}

\authorrunning{H. Jiang et al.}   
%

%
\institute{School of Robotics, XJTLU Entrepreneur College (Taicang)\\Xi'an Jiaotong-Liverpool University~\email{h.jiang@xjtlu.edu.cn}
\and Data Science Research Center\\ Duke Kunshan University~\email{guanyu.yang@dukekunshan.edu.cn}
\and Department of Communications and Networking , School of Advanced Technology\\Xi'an Jiaotong-Liverpool University~\email{fei.cheng@xjtlu.edu.cn}\\
\and Data Science Research Center\\Duke Kunshan University~\email{kaizhu.huang@dukekunshan.edu.cn}
}

\maketitle              

\begin{abstract}
Synthesizing Chinese characters with consistent style using few stylized examples is challenging. Existing models struggle to generate arbitrary style characters with limited examples. In this paper, we propose the \textbf{Generalized W-Net}, a novel class of W-shaped architectures that addresses this. By incorporating Adaptive Instance Normalization and introducing multi-content, our approach can synthesize Chinese characters in any desired style, even with limited examples. It handles seen and unseen styles during training and can generate new character contents. Experimental results demonstrate the effectiveness of our approach.
\end{abstract}


\section{Introduction}
\begin{figure}[htbp]
	\centering
	\subfigure[Characters with complex structures.]{
		\includegraphics[width=0.45\linewidth]{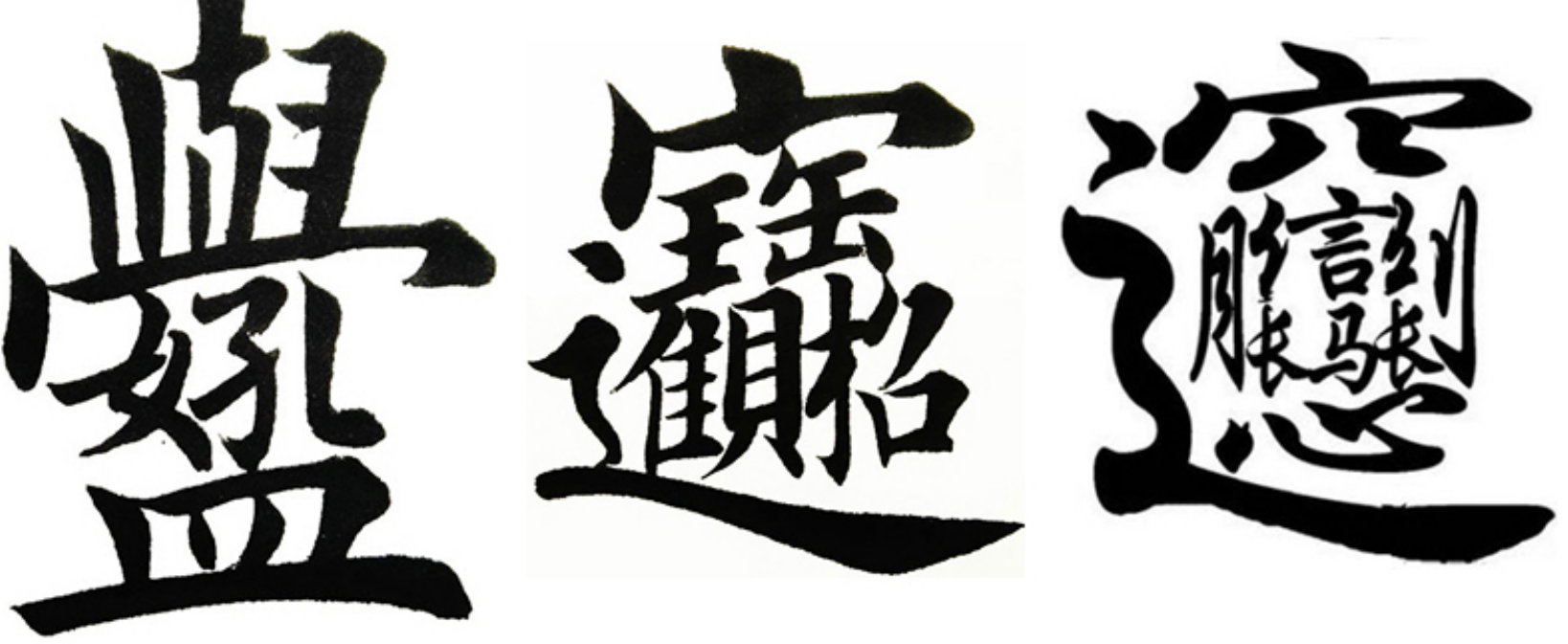}
		\label{SubFig:ComplicatedChars} }\quad
	\subfigure[A calligraphy by the Emperor Huizong of the Chinese Song Dynasty, Ji Zhao (中国宋朝徽宗皇帝赵佶), in the  1100s A.D.] {
		\includegraphics[width=0.45\linewidth]{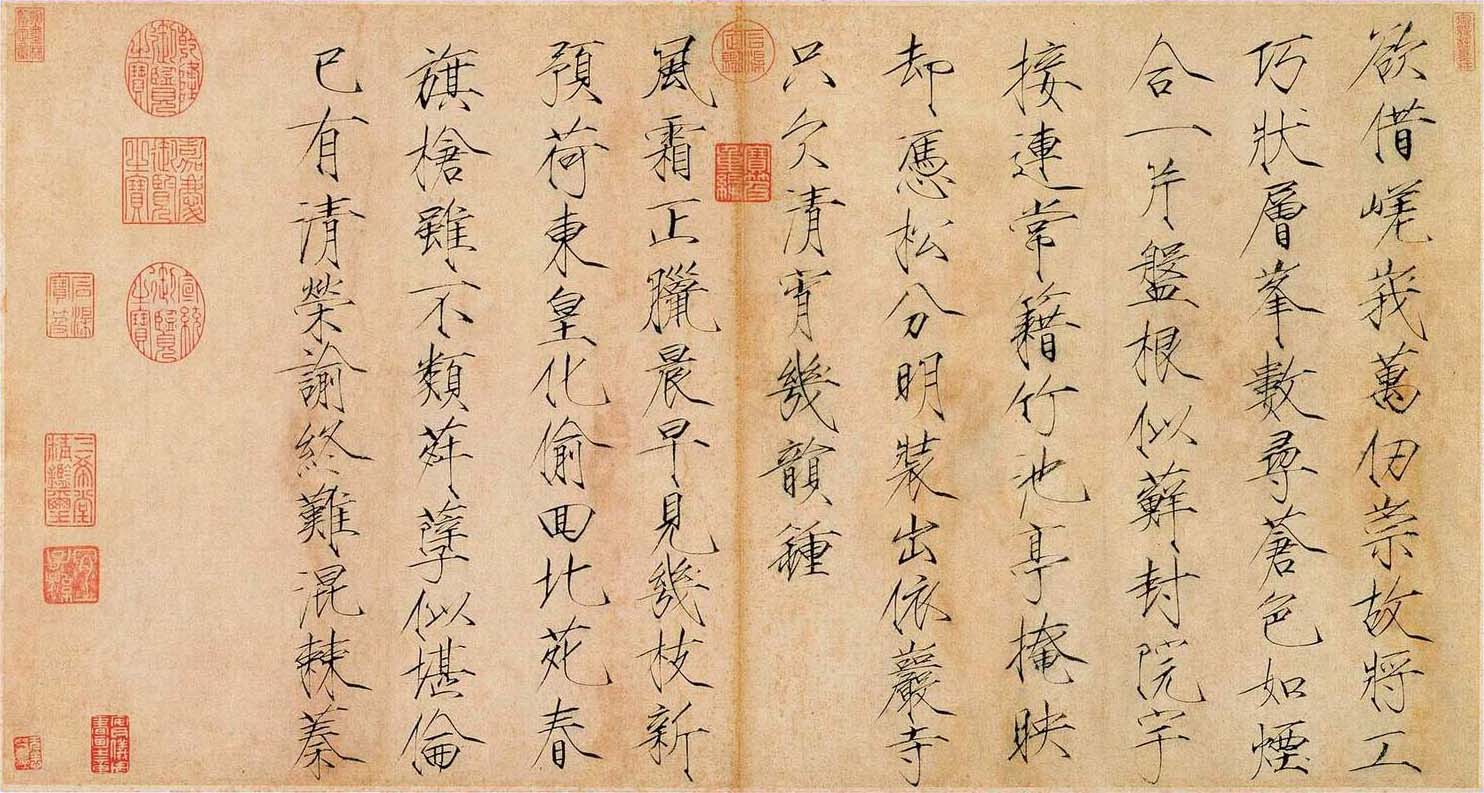}
		\label{SubFig:QianLongCalligraphy} }  
	\caption
	{Examples of Chinese characters and calligraphic work.}
	\label{Fig:ChineseExamples}
\end{figure}

Alphabet-based languages like English, German, and Arabic horizontally concatenate basic letters to form words. In contrast, oriental Asian languages such as Chinese, Korean, and Japanese use radicals and strokes as the minimum units for block characters. These units can vary horizontally and vertically, resulting in a diverse range of characters, as shown in Fig. \ref{SubFig:ComplicatedChars}, with over 50 strokes and 10 radicals per character. Handwritten block calligraphies possess not only messaging functions but also artistic and collectible value, as depicted in Fig. \ref{SubFig:QianLongCalligraphy}. This property is rarely seen in alphabet-based languages.

Synthesizing a large number of artistic and calligraphic characters in a consistent style, based on a few or even just one stylized example, is a highly challenging task~\cite{jiang2018w}. Previous attempts in few-shot character generation using deep generative neural networks have not proven effective on public Chinese handwritten datasets. While deep models have been successful in synthesizing vector fonts, they are less suitable for hand-written glyph generation, where image-based approaches are more appropriate.

\begin{figure}[htbp!]
	\centering
	\includegraphics[width=0.7\linewidth]{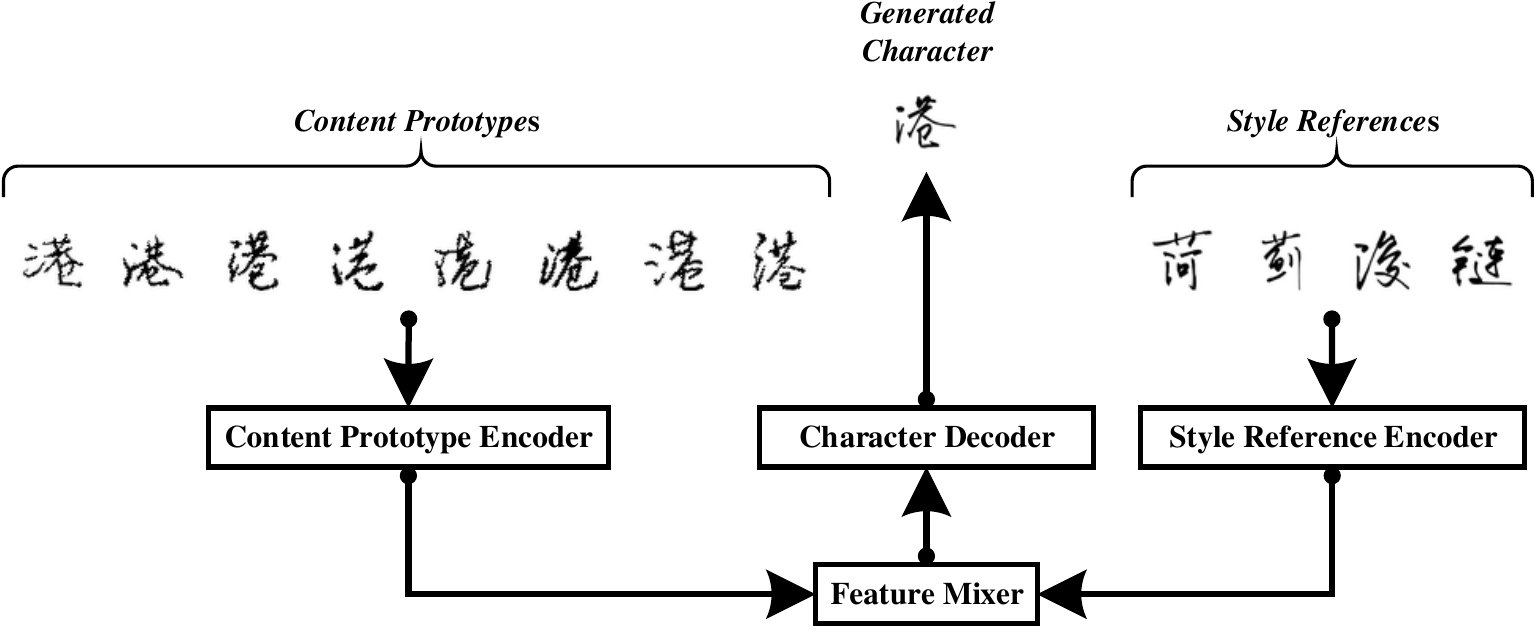}    
	\caption
	{Structure of W-shaped architectures for the character generation task.}\label{Fig:W-Shaped-Architecture}
\end{figure}

One successful approach, the \textit{W-Net}, employs a W-shaped architecture that effectively separates content and style inputs, specifically on public Chinese handwritten datasets~\cite{liu2011casia}. It consists of two parallel encoders: the style reference encoder and the content prototype encoder~\cite{jiang2018w, zhang2018unified}. The style reference encoder takes a set of stylized characters with different contents (referred to as \textit{style references}), while the content prototype encoder receives a set of characters with identical content but varying styles (known as \textit{content prototypes}). The extracted features from both encoders are combined using a feature mixer to generate the desired output. This W-shaped deep generative model, depicted in Fig. \ref{Fig:W-Shaped-Architecture}, leverages adversarial training~\cite{goodfellow2014generative, gulrajani2017improved} to synthesize a character that closely resembles the style of the \textit{style references} while preserving the content from the \textit{content prototypes}\footnote{The input of the \textit{W-Net} architecture in~\cite{jiang2018w} represents a special case, involving only one single content prototype with a standard font. In most cases, the styles are pre-selected and fixed prior to training or utilization.}.

In this paper, the proposed \textbf{Generalized W-Net} architecture incorporates residual blocks and dense blocks to enhance model performance. Various normalization techniques, such as batch normalization, instance normalization, layer normalization, and adaptive instance normalization (AdaIN), are employed to improve style transfer performance. Based on the \textit{W-Net} framework~\cite{jiang2018w}, the \textbf{Generalized W-Net} introduces multiple content prototypes with different styles, enabling the generation of characters that combine these prototypes non-linearly to achieve desired styles. Additionally, AdaIN is utilized in different parts of the framework to further enhance style transfer. Extensive experiments demonstrate the effectiveness and capability of the \textbf{Generalized W-Net} both visually and statistically.

\section{Preliminaries}
Notations and preliminaries necessary to demonstrate the proposed \textbf{Generalized W-Net} architecture are specified in this section.
Firstly, $X$ is denoted as a character dataset, consisting of $J$ different characters with in total $I$ different fonts.
\begin{defn}\label{Def:WNet_RealTarget}
	Let $x^i_j$ be a specific sample in  $X$, regarded as the \textit{real target}.
	Following~\cite{jiang2017field}, the superscript $i \in [1,2,...,I]$ represents $i$-th character style, while the subscript $j \in [1,2,...,J]$ denotes the $j$-th character content.
\end{defn}
\begin{defn}\label{Def:WNet_ContentPrototype}
	Denote $x^{c_m}_j, c_m \in [1,2,...,I], m=1,2,...,M$ be the set of \textit{content prototype}s.
	It describes a content of the $j$-th character, the same as the content of $x^i_j$ defined in Definition~\ref{Def:WNet_RealTarget}.
\end{defn}
Particularly, $M$ different styles are to be pre-selected before the model is optimized. 
Commonly, they are out of the fonts for the real target ($[1,2,...,I]$). 
\begin{defn}\label{Def:WNet_StyleReference}
	Denote $x^i_{s_n}, s_n \in [1,2,...,J], n=1,2,...,N$ be a set of \textit{style reference}s with the $i$-th style, identical to the style of $x^i_j$ defined in Definition~\ref{Def:WNet_RealTarget}.
\end{defn}
Note that $i$ and $c_m$ are generally different, while $j$ and $s_n$ also differ. The number of \textit{style references} $N$ should be determined before training, but can vary during testing\footnote{In the proposed \textbf{Generalized W-Net} and \textit{W-Net} architecture~\cite{jiang2018w}, $N$ can be changed during testing.}. In our model, each $x^i_j$ combines the $j$-th content information from $x^{c_m}j$ prototypes and the $i$-th writing style learned from $x^i{s_n}$ references. Here, $m\in[1,2,...,M]$ and $n\in[1,2,...,N]$ follow the definitions in Definition~\ref{Def:WNet_ContentPrototype} and Definition~\ref{Def:WNet_StyleReference}, respectively.
\begin{defn}
	The proposed \textbf{Generalized W-Net} model will synthesize the \textit{generated character} by taking \textit{content prototype}s ($x^{c_1}_j,x^{c_2}_j,...,x^{c_M}_j$, as defined in Definition~\ref{Def:WNet_ContentPrototype}) and \textit{style reference}s ($x^i_{s_1},x^i_{s_2},...,x^i_{s_N}$, as defined in Definition~\ref{Def:WNet_StyleReference}) simultaneously.
	The corresponding \textit{generated character} is denoted as $G(x^{c_1}_j,x^{c_2}_j,...,x^{c_M}_j, x^i_{s_1},x^i_{s_2},...,x^i_{s_N})$\footnote{The \textit{generated character} will be noted as $G(x^{c_m}_j,x^i_{s_n})$ for simplicity.}.
\end{defn}

The objective of training is to make the generated character $G(x^{c_m}j,x^i{s_n})$ resemble the real target $x^i_j$ in content and style. In the few-shot setting, a set of content prototypes ($x^{c_m}q, m=1,2,...,M$) is combined with the style reference encoder's outputs ($Enc_r(x^h{p_1},x^h_{p_2},...,x^h_{p_L})$) to produce characters ($G(x^{c_m}q, x^h{p_l})$) with the desired style. Each value of $q$ generates a different character that imitates the given styles ($x^h_{p_l}, l=1,2,...,L$). The values of $p_l$ and $q$ can vary within specific ranges, and $h$ can be outside certain ranges to produce the desired character variations given a few styles~\footnote{The output of the style reference encoder $Enc_r(x^h_{p_1},x^h_{p_2},...,x^h_{p_L})$ is connected to the $Dec$ using shortcut or residual/dense block connections (see Section~\ref{Sec:FeatureMixer}).}.

\section{The Generalized W-Net Architecture}
\begin{figure*}[hbtp!]
	\centering
	\includegraphics[width=\linewidth]{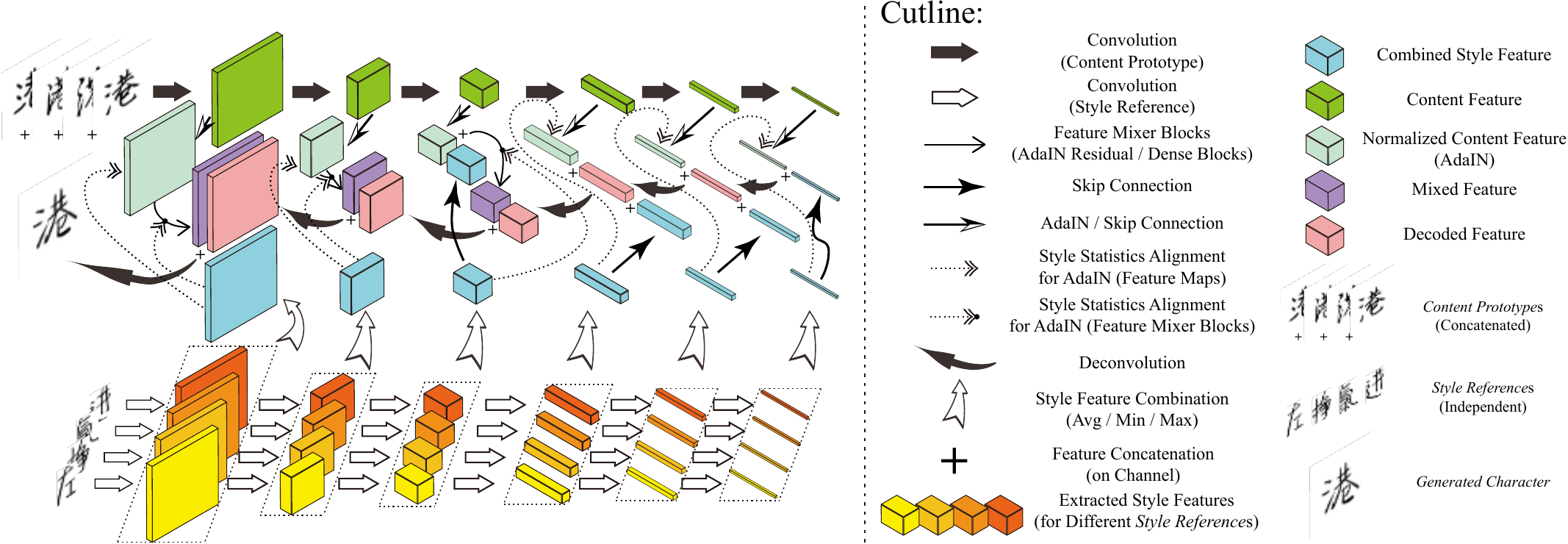}    
	\caption
	{Model Architecture: \textbf{Generalized W-Net}}\label{Fig:Upgraded-W-Net}
\end{figure*}

Fig.~\ref{Fig:Upgraded-W-Net} illustrates the structure of the proposed \textbf{Generalized W-Net}. It includes the content prototype encoder ($Enc_p$), the style reference encoder ($Enc_r$), the feature mixer, and the decoder ($Dec$).

\subsection{Encoders}
The $Enc_p$ and $Enc_r$ are formed using convolutional blocks, each comprising convolutional filters ($5\times5$ kernel size with stride 2), normalization\footnote{The specific normalization method may vary across implementations (see Section~\ref{Sec:FeatureMixer-AdaIN}).}, and ReLU activation. With this configuration, $M$ $64\times64$ prototypes $x^{c_m}_j$ ($c_m \in [1,2,...,I], m =1,2,...,M$) and $N$ references $x^i_{s_n}$ ($s_n \in [1,2,...,J], n =1,2,...,N$) are both mapped to $1\times512$ feature vectors: $Enc_p(x^{c_1}_j,x^{c_2}_j,...,x^{c_M}_j)$ and $Enc_r(x^i_{s_1},x^i_{s_2},...,x^i_{s_N})$ respectively.\footnote{The encoded outputs will be referred to as $Enc_p(x^{c_m}_j)$ and $Enc_r(x^i_{s_n})$.}

The single-channel \textit{content prototypes} are concatenated channel-wise, resulting in an $M$-channel input for $Enc_p$. The hyper-parameter $M$ remains fixed during training and real-world application. On the other hand, $N$ input \textit{style references} generate $N$ features using shared weights from $Enc_r$. The combined style features are obtained by averaging, maximizing, and minimizing these $N$ style features. They are then concatenated channel-wise to produce $Enc_r(x^i_{s_n})$. Therefore, the choice of $N$ (or $L$ during testing) can vary due to weight sharing.


\begin{figure}[htbp]
	\centering
	\subfigure[Feature Mixer with BN]{
		\includegraphics[width=0.45\linewidth]{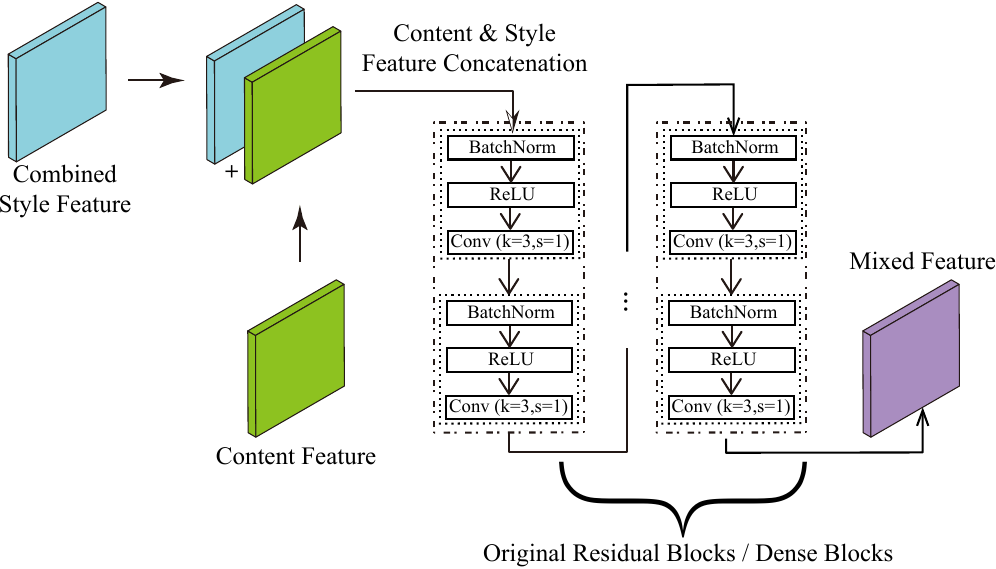}
		\label{Fig:BatchNorm-FeatureMixer} }
	\subfigure[Feature Mixer with AdaIN] {
		\includegraphics[width=0.45\linewidth]{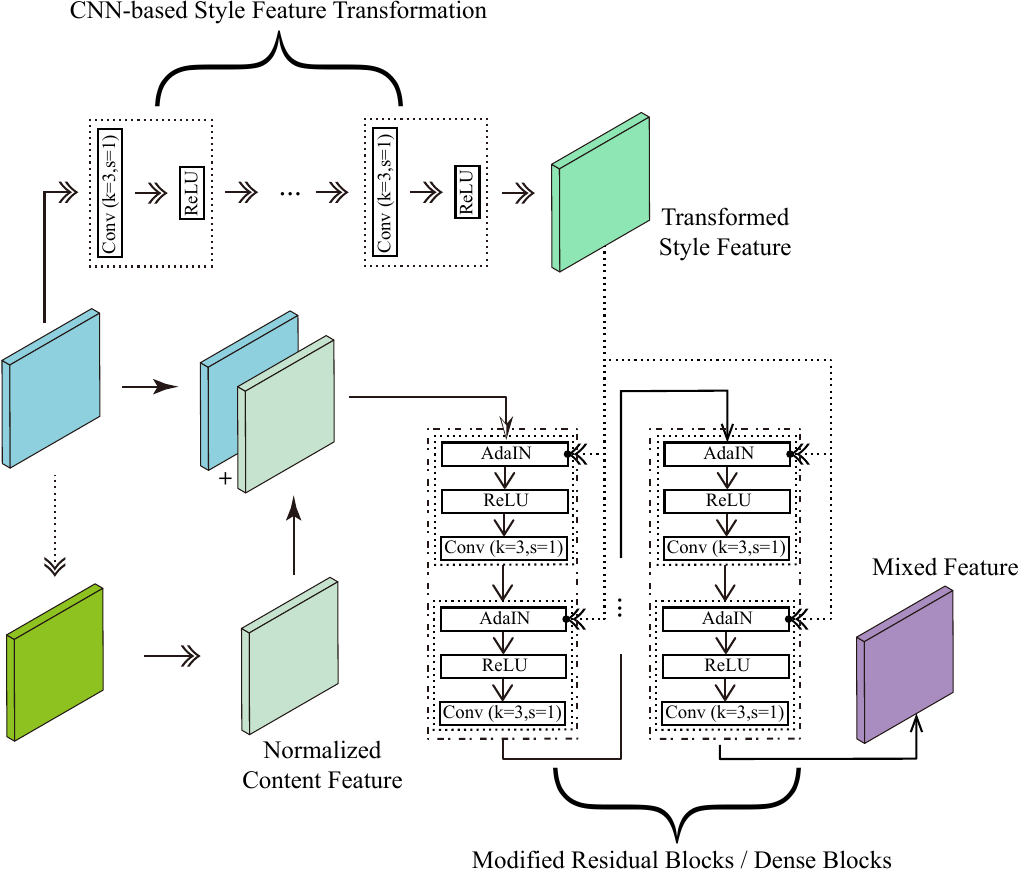}
		\label{Fig:AdaIN-FeatureMixer} }  
	\caption
	{Examples of Chinese characters and calligraphic work.}
	\label{Fig:FixerMixer}
\end{figure}

\subsection{Feature Mixer}\label{Sec:FeatureMixer}
Taking inspiration from the U-Net architecture~\cite{jiang2018w, jiang2018style}, the encoder and character decoder feature mutual connections between equivalent layers for combining content and style. These connections, discussed in Section~\ref{Sec:Decoder}, involve a feature mixer that incorporates batch normalization (BN). An upgraded version, Adaptive Instance Normalization (AdaIN), will be explained in Section~\ref{Sec:FeatureMixer-AdaIN}.

Feature maps of deeper layers' from both encoders are concatenated channel-wise, while only the content feature is used in shallower levels, resulting in combined features. In shallower layers, these combined features are processed through multiple dense blocks~\cite{huang2017densely} or residual blocks~\cite{he2016identity} for further enhancement. Each block consists of batch normalization, ReLU activation, and convolutional filters, as shown in Fig.\ref{Fig:BatchNorm-FeatureMixer}. The concatenated feature in deeper layers serves as the mixed feature, simplifying to the original shortcut connection used in the U-Net architecture\cite{jiang2018style}.

\subsection{AdaIN in the Feature Mixer}\label{Sec:FeatureMixer-AdaIN}
The AdaIN~\cite{huang2017arbitrary} has been proven effective for enhancing style transfer performance in various relevant research studies, such as those mentioned in~\cite{zhang2018unified,zhang2018separating,huang2018multimodal}. It is defined by Eq.(\ref{Eqt:AdaIN}), where the statistics $\sigma(\beta)$ and $\mu(\beta)$ of the style features are computed across spatial locations to normalize the content feature $\alpha$. The calculations for $\sigma(\beta)$ and $\mu(\beta)$ are given by Eqs.(\ref{Eqt:AdaIN_Sigma}) and (\ref{Eqt:AdaIN_Mu}), respectively.
\begin{equation}\label{Eqt:AdaIN}
	\begin{aligned}
		\textbf{AdaIN}(\alpha,\beta)=\sigma(\beta)\cdot(\frac{\alpha-\mu(\alpha)}{\sigma(\alpha)})+\mu(\beta)
	\end{aligned}
\end{equation}
\begin{equation}\label{Eqt:AdaIN_Sigma}
	\begin{aligned}
		\sigma_{nc}(\beta)=\sqrt{\frac{1}{HW}\sum_{h,w}^{H,W}(\beta_{nchw}-\mu_{nc}(x))^2+\epsilon}
	\end{aligned}
\end{equation}
\begin{equation}\label{Eqt:AdaIN_Mu}
	\begin{aligned}
		\mu_{nc}(\beta)=\frac{1}{HW}\sum_{h,w}^{H,W}\beta_{nchw}
	\end{aligned}
\end{equation}
$N$, $C$, $H$, and $W$ represent the batch size, number of channels, spatial height, and spatial width of the style feature map $\beta$. It signifies the statistical alignment normalization between a content feature map $\alpha$ and the style feature map $\beta$. In the proposed feature mixer of the \textbf{Generalized W-Net}, AdaIN can be positioned in two locations, as shown in Fig.~\ref{Fig:AdaIN-FeatureMixer}.

\subsubsection{AdaIN on the Extracted Content Feature}
The extracted content features ($f^\gamma_p(x^{c_m}_j)$) in the $\gamma$-th layer of the content encoder are normalized by the statistics of the combined style feature ($f^\gamma_r(x^i_{s_n})$) on the equivalent layer of the style encoder.
It is performed by Eqt.~(\ref{Eqt:AdaIN}), Eqt.~(\ref{Eqt:AdaIN_Sigma}), and Eqt.~(\ref{Eqt:AdaIN_Mu}), where $\alpha=f^\gamma_p(x^{c_m}_j)$, $\beta=f^\gamma_r(x^i_{s_n})$, $H=H_\gamma$, $W=W_\gamma$ ($H_\gamma$ and $W_\gamma$ are the corresponding feature sizes on the $\gamma$-th layer).

\subsubsection{AdaIN in the Dense / Residual Blocks}
BNs in the residual/dense blocks are replaced with AdaINs. Inspired by~\cite{huang2018multimodal}, multiple Convolutional Neural Network (CNN)-based $Conv(3\times3)-ReLU$ transformations process the combined style feature, resulting in the transformed style feature $g(f^\gamma_r(x^i_{s_n}))$ at the $\gamma$-th layer. Within these blocks ($h$), AdaIN replaces BN, as specified by Eq.(\ref{Eqt:AdaIN}), Eq.(\ref{Eqt:AdaIN_Sigma}), and Eq.~(\ref{Eqt:AdaIN_Mu}), using the statistics of the transformed style feature $g(f^\gamma_r(x^i_{s_n}))$ instead of the style feature $f^\gamma_r(x^i_{s_n})$.

\subsubsection{Other Corporations}
When AdaIN is not used, both the encoders and decoder utilize Batch Normalization (BN). However, as proposed in~\cite{huang2018multimodal}, when AdaIN is employed, BN in the style encoder is omitted. Additionally, in the content encoder, Layer Normalization (LN) is used, and in the decoder, Instance Normalization (IN) is used. The summary of these settings can be found in Table~\ref{Tab:Normalization}.
\begin{table}[htbp!]
	\centering
	\tiny
	\caption{Normalization utilized in the \textbf{Generalized W-Net} Architecture}\label{Tab:Normalization}
	\begin{tabular}{|c|c|c|c|}
		\hline
		\multicolumn{2}{|c|}{}                                   & BN-based & AdaIN-based \\ \hline
		\multicolumn{2}{|c|}{Content Encoder}                    & BN       & LN          \\ \hline
		\multicolumn{2}{|c|}{Style Encoder}                      & BN       & N/A         \\ \hline
		\multirow{2}{*}{Feature Mixer} & Content Feature Alignment         & N/A      & AdaIN       \\ \cline{2-4} 
		& Residual / Dense Blocks & BN       & AdaIN       \\ \hline
		\multicolumn{2}{|c|}{Character Decoder}                  & BN       & IN          \\ \hline
	\end{tabular}
\end{table}

\subsection{Character Decoder}\label{Sec:Decoder}
The $Dec$ in the \textbf{Generalized W-Net} is designed to be similar to the decoder in the \textit{W-Net}\cite{jiang2018w}. It consists of deconvolutional blocks connected to the mixed feature. Each block in the $Dec$ includes deconvolutional filters (kernel size: $5\times5$, stride: 2), normalization, and a non-linear activation function. The Leaky ReLU is used for all layers except the last one\cite{radford2015unsupervised}. The final output, referred to as the \textit{generated character}~\cite{goodfellow2014generative}, is obtained by applying the $tanh$ nonlinearity.

\subsection{Training Strategy and Optimization Losses}\label{Sec:Losses}
The proposed \textbf{Generalized W-Net}\cite{jiang2018w} is trained using the Wasserstein Generative Adversarial Network with Gradient Penalty (W-GAN-GP) framework\cite{gulrajani2017improved}. It serves as the generative model, denoted as $G$, taking content prototypes and style references as inputs to generate characters. The generation process is represented by $G(x^{c_m}j, x^i{s_n})=Dec(Enc_p(x^{c_m}j), Enc_r(x^i{s_n}))$, where $Dec$ and $Enc$ refer to the decoder and encoder, respectively. The objective is to optimize this formulation to make it similar to $x_j^i$ using reconstruction, perceptual, and adversarial losses, determined by the discriminative model $D$ within the WGAN-GP framework.

\subsubsection{Training Strategy} 
In each alternative training iteration, $G$ and $D$ are optimized in an alternative manner.
$G$ optimizes $\mathbb{L}_{G}=-\alpha\mathbb{L}_{adv-G}+\beta\mathbb{L}_{ac}
+\lambda_{pixel}\mathbb{L}_{pixel}+\mathbb{L}_{\phi_{total}}+\psi_{p}\mathbb{L}_{Const_p}+\psi_{r}\mathbb{L}_{Const_r}$. Simultaneously, $D$ minimizes  $\mathbb{L}_{D}=\alpha\mathbb{L}_{adv-D}+\alpha_{GP}\mathbb{L}_{adv-GP}+\beta\mathbb{L}_{ac}$.

\subsubsection{Adversarial Loss}
The adversarial losses are given by $\mathbb{L}_{adv-G} = D(x^{c_{m'}}_j, G(x^{c_m}_j,x^i_{s_n}),x^i_{s_{n'}})$ and  $\mathbb{L}_{adv-D} = D(x^{c_{m'}}_j, x^i_j, x^i_{s_{n'}})  - D(x^{c_{m'}}_j, G(x^{c_m}_j,x^i_{s_n}),x^i_{s_{n'}})$ for $G$ and  $D$ respectively. 
$m'$ and $n'$ are randomly sampled from $[1,2,...,M]$ and $[1,2,...,N]$ respectively for each training example $x^i_j$. 
The gradient penalty $\mathbb{L}_{adv-GP} = ||\nabla_{\widehat{x}}D(x^{c_{m'}}_j, \widehat{x},x^i_{s_{n'}})-1||_2$~\cite{gulrajani2017improved} is an essential part to make $D$ satisfy the Lipschitz continuity condition required by the Wasserstein-based adversarial training.
$\widehat{x}$ is uniformly interpolated along the line between  $x^i_j$ and $G(x^{c_m}_j,x^i_{s_n})$.

\subsubsection{Categorical Loss of the Auxiliary Classifier}
As notified in~\cite{odena2016conditional}, the auxiliary classifier on the discriminator is optimized by $\mathbb{L}_{ac} = \left[\log C_{ac}(i|x^{c_{m'}}_j, x^i_j, x^i_{s_{n'}})\right] + \left[\log C_{ac}(i|x^{c_{m'}}_j, G(x^{c_m}_j,x^i_{s_n}),x^i_{s_{n'}})\right]$.

\subsubsection{Constant Losses of the Encoders}
The constant losses~\cite{taigman2016unsupervised} are employed to better optimize both encoders. 
They are given by $\mathbb{L}_{Const_p} = ||Enc_p(x^{c_m}_j)-Enc_p(G(x^{c_m}_j,x^i_{s_n}))||^2$ and $\mathbb{L}_{Const_r} = ||Enc_r(x^i_{s_n})-Enc_r(G(x^{c_m}_j,x^i_{s_n}))||^2$ respectively for $Enc_p$ and $Enc_r$.

\subsubsection{Pixel Reconstruction Losses}
It represents the discrepancy of pixels from two comparing images, namely, the \textit{generated character} and the corresponding ground truth
It is specified by the L1 difference $\mathbb{L}_{pixel} = ||(x^i_j-G(x^{c_m}_j,x^i_{s_n}))||_1$. 

\subsubsection{Deep Perceptual Losses}
The deep perceptual loss aims to minimize the variation between the generated character and the corresponding input images by calculating differences in high-level features. The total perceptual loss, denoted as $\mathbb{L}{\phi{total}}$, is composed of three components: $\mathbb{L}{\phi{real}}$, $\mathbb{L}{\phi{content}}$, and $\mathbb{L}{\phi{style}}$, each incorporating mean square error (MSE) discrepancy and von-Neumann divergence~\cite{yang2018learning}.

The loss function $\mathbb{L}{\phi{real}}$ compares the generated character $G(x^{c_m}j,x^i{s_n})$ with the corresponding input $x^i_j$. It utilizes a pre-trained VGG-16 network, denoted as $\phi_{real}$, which is used for character content and writing style classification. Different convolutional feature variations, $\phi_{1-2}$, $\phi_{2-2}$, $\phi_{3-3}$, $\phi_{4-3}$, and $\phi_{5-3}$, are considered, representing features from specific layers and blocks in the VGG-16 Network.

For $\mathbb{L}{\phi{content}}$, the goal is to differentiate the generated character $G(x^{c_m}j,x^i{s_n})$ in style while preserving the same character content $x^{c_{m'}}j$. It relies on the $\phi{content}$ network, trained exclusively for character content classification. To minimize differences in high-level abstract features, only the $\phi_{4-3}$ and $\phi_{5-3}$ features are considered.

Similarly, $\mathbb{L}{\phi{style}}$ aims to differentiate the generated character $G(x^{c_m}j,x^i{s_n})$ in character content while preserving the same writing style $x^i_{s_{n'}}$. The $\phi_{style}$ network is trained by minimizing cross-entropy for accurate style classification. Again, only the $\phi_{4-3}$ and $\phi_{5-3}$ features are utilized to ensure similarity in high-level patterns.

\section{Experiments}\label{Sec:Exp}
We only report a few visual examples that are able to demonstrate the effectiveness of the proposed \textbf{Generalized W-Net} in this section. Detailed objective results will be in the future work.
\begin{figure}[h!]
	\centering
	\includegraphics[width=0.6\linewidth]{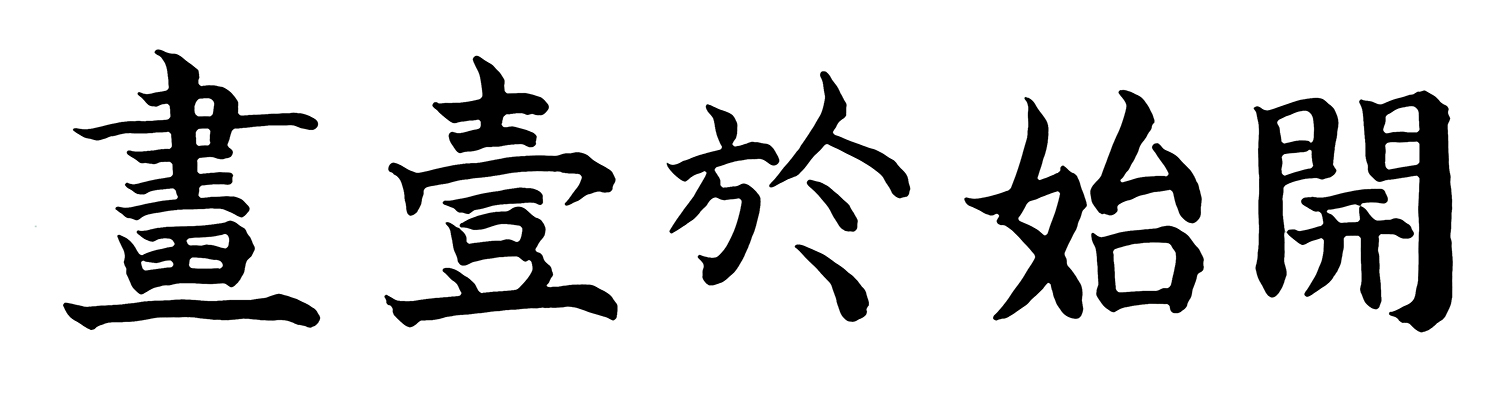}    
	\caption
	[The input few brush-written examples of actual Chinese characters]
	{The input few brush-written examples of actual Chinese characters.\footnotemark}\label{Fig:WNet_FurtherStudies_InputFew}
\end{figure}
\footnotetext{These brush-written simplified Chinese character are written by Dr. Fei CHENG from School of Advanced Technology, Xi'an Jiaotong-Liverpool University.}

\begin{figure}[h!]
	\centering
	\subfigure[Generated Chinese]{
		\includegraphics[width=0.31\linewidth]{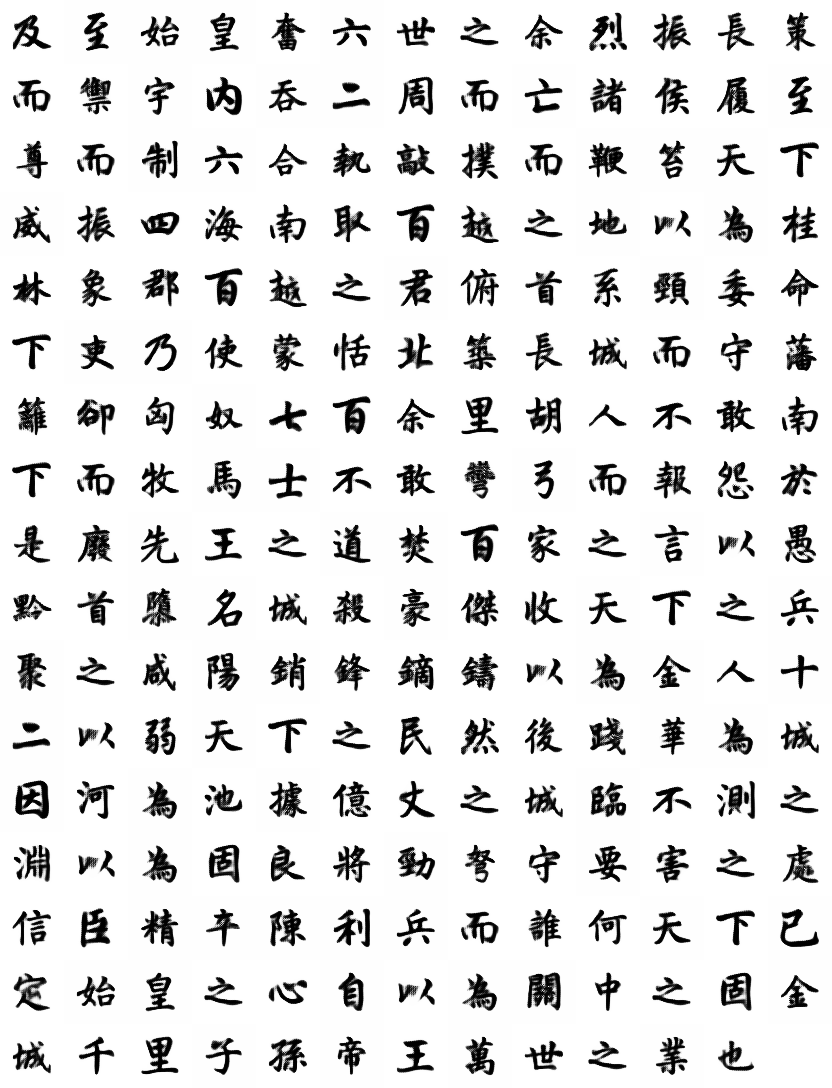}
		\label{Fig:GeneratedTraditionalChinese} }
	\subfigure[Generated Korean] 
	{\includegraphics[width=0.31\linewidth]{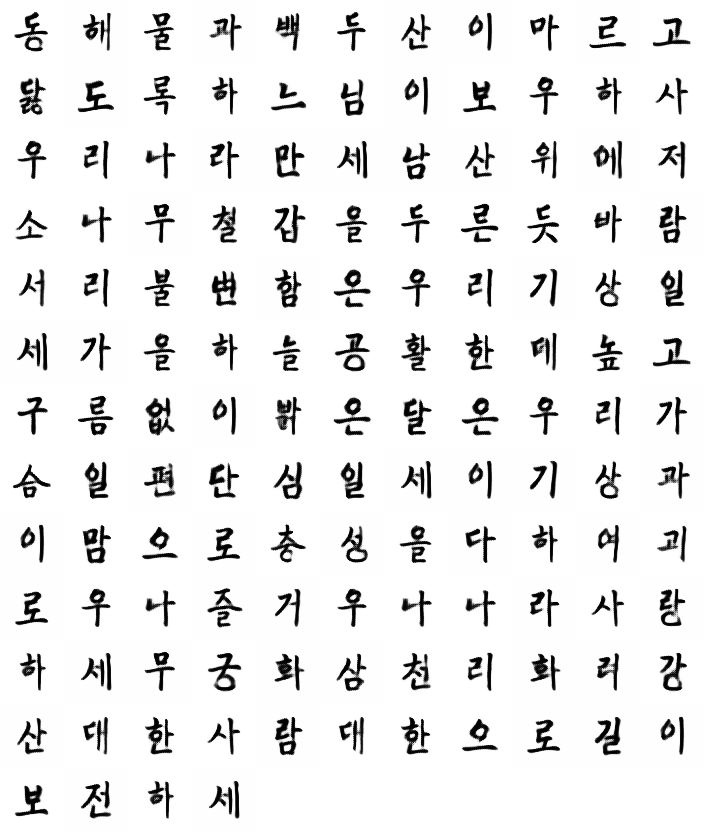}
		\label{Fig:GeneratedKorean} }	
\subfigure[Generated Japanese]{
		\includegraphics[width=0.31\linewidth]{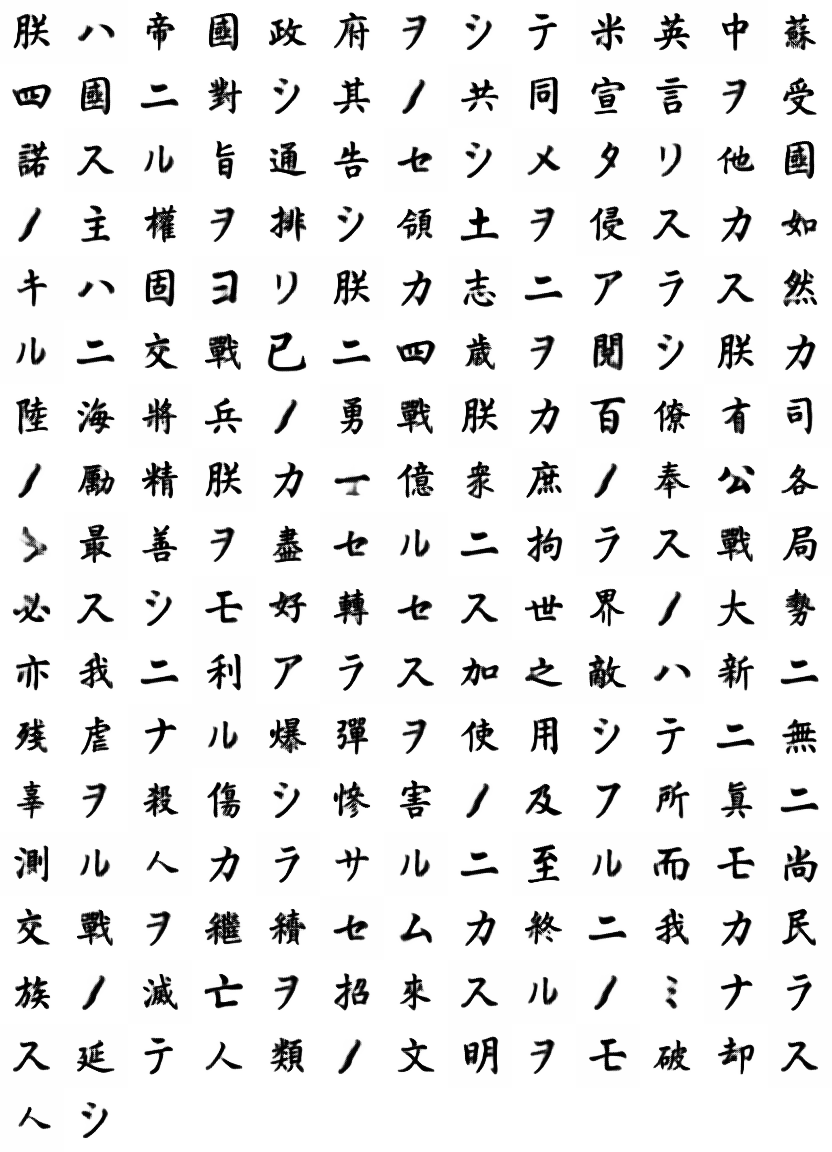}
		\label{Fig:GeneratedJapanese} }
	\caption[]
	{The W-Net (trained with only simplified Chinese characters) generated essay of traditional Chinese, Korean, and Japanese characters from the one selected style reference shown in Fig.~\ref{Fig:WNet_FurtherStudies_InputFew}}
	\label{Fig:ChnJpnKrn}
\end{figure}

Most of the characters in the eastern Asian languages including Chinese (traditional or simplified), Korean, and Japanese are constructed by rectangular shapes (known as the \emph{block characters}~\cite{chinese_character}).
In this sense, it is an interesting evaluation to make the well-optimized W-Net model to generate characters of these three kinds of languages. 
The training of the the proposed \textit{Generalized W-Net} follows the description in Section~\ref{Sec:Losses} where only the simplified Chinese characters are available.
For each of the generation processes of paragraphs in these kinds of languages, one randomly selected brush-written character listed in Fig.~\ref{Fig:WNet_FurtherStudies_InputFew} will be specified as the one-shot style reference. 
Fig.~\ref{Fig:ChnJpnKrn} illustrates the generated result of the corresponding a traditional Chinese poetry, a Korean lyric, and a Japanese speech.  
It can be obviously found that the style tendency given in Fig.~\ref{Fig:WNet_FurtherStudies_InputFew} is well kept in all the three generated outputs. 

\begin{figure}[h!]
	\centering
	\subfigure[Content Prototypes]{
		\includegraphics[width=0.35\linewidth]{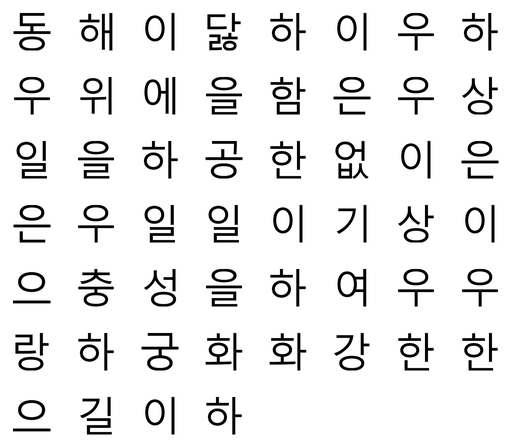}
		\label{SubFig:KoreanCircleSource} }\quad\quad
	\subfigure[Generated Characters] 
	{\includegraphics[width=0.35\linewidth]{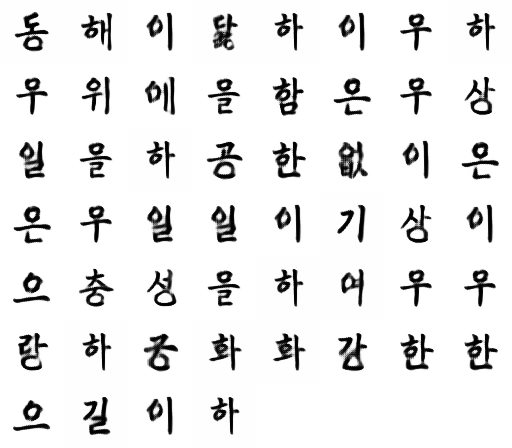}
		\label{SubFig:KoreanCircleTarget} }	
	\caption[Some Korean characters with \emph{circular radical}s]
	{Some Korean characters with \emph{circular radical}s. The characters with circular radicals are selected from the ones given in Fig.~\ref{Fig:GeneratedKorean}}
	\label{Fig:WNet_KoreanCircle}
\end{figure}

A more interesting finding is that the proposed \textbf{Generalized W-Net} model trained with simplified Chinese characters is always capable to synthesize the \emph{circular radical}s that are commonly seen in Korean but rarely found in Chinese.
As seen in Fig.~\ref{Fig:WNet_KoreanCircle}, the \emph{circular radical}s are mostly preserved and recovered in the generated characters (Fig.~\ref{SubFig:KoreanCircleTarget}) when compared with the corresponding content prototypes shown in Fig.~\ref{SubFig:KoreanCircleSource}.
It further demonstrates that the proposed W-Net is capable of being generalized to the useful knowledge in strokes and radicals that are absent from the training data.

\section{Conclusion}
A novel generalized framework \textbf{Generalized W-Net} is introduced in this paperto achieve Few-shot Multi-content Arbitrary-style Chinese Character Generation (FMACCG) task.
Specifically, the proposed model, composing of two encoders, one decoder, and a feature mixer with several layer-wised connections, is trained adversarially based on the Wasserstein GAN scheme with the gradient penalty.
It enables synthesizing any arbitrary stylistic character by transferring the learned style information from one single style reference to the input single content prototype.
Extensive experiments have demonstrated the reasonableness and effectiveness of the proposed \textbf{Generalized W-Net} model in the few-shot setting.

%
%
\bibliographystyle{splncs04}



%
\end{CJK*}
\end{sloppypar}
\end{document}